\documentclass{article} 
\usepackage{iclr2025_conference,times}

\usepackage{amsfonts}
\usepackage{amsmath}
\usepackage{booktabs}
\usepackage{graphicx}
\usepackage{multirow,multicol}
\usepackage{subcaption}
\usepackage{hyperref}
\usepackage{url}
\usepackage{xspace}
\usepackage{xcolor}

\newcommand{\IR}{\mathbb{R}}
\newcommand{\V}{\mathcal{V}}
\newcommand{\E}{\mathcal{E}}

\definecolor{mulberry}{rgb}{0.77, 0.29, 0.55}

\newcommand{\comment}[1]{}

\newcounter{ObservationCounter}

\newif\ifsubmit
\submittrue 
\ifsubmit
    \newcommand{\isaiah}[1]{}
    \newcommand{\ramiro}[1]{}
    \newcommand{\ben}[1]{}
    \newcommand{\howie}[1]{}
    \newcommand{\outline}[2]{}
    
    \newcommand{\rev}[2]{}
\else
    \newcommand{\isaiah}[1]{\textcolor{mulberry}{\textbf{Isaiah: #1}}}
    \newcommand{\ramiro}[1]{\textcolor{blue}{\textbf{Ramiro: #1}}}
    \newcommand{\ben}[1]{\textcolor{purple}{\textbf{Ben: #1}}}
    \newcommand{\howie}[1]{\textcolor{red}{\textbf{Howie: #1}}}
    \newcommand{\outline}[2]{\textcolor{blue}{\##1:~}\textcolor{cyan}{#2}}
    
    \newcommand{\rev}[2]{\textcolor{green}{Review #1: ``#2"}}
\fi

\title{Automated Cyber Defense with Generalizable Graph-based Reinforcement Learning Agents}

\author{Isaiah J. King$^{1,2}$, Benjamin Bowman$^1$, \& H. Howie Huang$^{1,2}$ \\
$^1$Cybermonic LLC, $^2$ The George Washington University GraphLab \\
\texttt{\{isaiah, benjamin, howie\}@cybermonic.com}
}

\iclrfinalcopy 
\begin{document}

\maketitle
\begin{abstract}
Deep reinforcement learning (RL) is emerging as a viable strategy for automated cyber defense (ACD). 
The traditional RL approach represents networks as a list of computers in various states of safety or threat. 
Unfortunately, these models are forced to overfit to specific network topologies, rendering them ineffective when faced with even small environmental perturbations. 
In this work, we frame ACD as a two-player context-based partially observable Markov decision problem with observations represented as attributed graphs.
This approach allows our agents to reason through the lens of relational inductive bias. 
Agents learn how to reason about hosts interacting with other system entities in a more general manner, and their actions are understood as edits to the graph representing the environment. 
By introducing this bias, we will show that our agents can better reason about the states of networks, and zero-shot adapt to new ones. 
We show that this approach outperforms the state-of-the-art by a wide margin, and makes our agents capable of defending never-before-seen networks against a wide range of adversaries in a variety of complex, and multi-agent environments. 
\isaiah{Abstract deadline: May 11; Full paper deadline: May 15; page limit: 9 (not including refs/appendicies)}
\end{abstract}

    \section{Introduction}
\label{sec:intro}

Automated cyber defense (ACD) systems are agents which, with no human intervention, defend a network from complex cyber-attacks--automated, or human~\citep{acd_review_paper}.
Like an autonomous security operations center, ACD agents monitor the network at all times, waiting to respond to a cyber-attack. 
When an incident occurs, the ACD agent would have the power to update firewall rules, reset machines, etc., to prevent the intruder from spreading. 
Using a simulated network environment, we can train agents for this task using reinforcement learning (RL). 
Prior works in this field compress the environment into a vector and feed it into deep RL networks~\citep{ddos_prevention_dqn,sdn_dqn,attack_attempts_viewable,attacks_and_defenses,simple_experiment}. 
This naive approach produces agents that are adept at defending the network they were trained on, but overly sensitive to minor changes in the network topology. 
As such, these agents are rarely evaluated in new and/or modified environments, a situation that is essential for any technology to be viable in the real world.

This observation coincides with a very old problem in RL: environmental overfitting~\citep{use_multi_envs}. 
In traditional supervised learning, a model has become overfit if it memorizes every data point and is unable to generalize about data it has never seen. 
To account for this, one would partition the data into disjoint training and testing sets to ensure the model can adapt to uncertainty. 
In RL, the uncertainty lies in how an action will affect the environment.
Environmental overfitting occurs when an agent no longer has uncertainty about the world it interacts with. 
Admittedly, model overfitting is not a problem in some cases; most RL benchmarks train and evaluate on the same environments~\citep{zsg_survey}, and why shouldn’t they?
An agent that plays chess can always expect the game to start with the same two sets of pieces placed in the same way across the same sixty-four squares. 
There is no reason to expect a chess-playing agent to adapt to a 9$\times$9 board with five knights.
But, for many real-world systems, such as ACD, the assumption that the environment will never change does not hold~\citep{graphs_as_states}.

One way to represent differences between networks' topologies and encourage agents to learn policies that generalize across environments is to represent them as graphs. 
The use of graphs for state representation has been done by several prior works in this field~\citep{simple_experiment,acd_g,crown_jewels}, but none process the graph directly.
Agents in prior work receive graphs representing the environment as part of their observations, but when processing them for their agents, the graph is compressed into a vector, and information is lost. 
Most often, the topological structure of the graph is thrown out entirely, and information about individual nodes is concatenated together~\citep{attacks_and_defenses,simple_experiment,booker2020model,cage1_hppo,cyber_battle_sim_decoy_experiment,keeping_it_rl}. 
This results in models that learn about implicit relationships between nodes but cannot adapt to explicit changes of novel network topologies because these changes cannot be communicated to the agents. 
Perturbations as simple as changing the order in which nodes are indexed can cause these models to perform no better than random.


In this work, we propose a novel, graph-centric strategy for highly generalizable automated network defense agents. 
Harnessing the power of relational inductive bias~\citep{gnn_motivation}, we provide our models with the full graph of the network, which they process without compression using a graph neural network (GNN)~\citep{gcn}. 
Inductive GNNs process graph input in a permutation-invariant manner and are structurally unable to rely on fixed node identities or positions.
Automorphisms between nodes are implicitly understood to GNN-based models as features and edges are identical regardless of node mapping. 
However, this is not the case with traditional models as the columns of their parameters are fixed, such that automorphic node permutations do not produce identical outputs.  
The output of this process is a matrix of node representations, called embeddings, which contain information about each node's features, as well as the features of their $k$-hop neighborhood. 
The node embeddings are then used as inputs to a policy network that selects the best action. 
Importantly, actions are not formulated as a fixed-length list, as is done by prior works, including graph-based approaches~\citep{globalnode}; instead, we represent actions as functions upon individual nodes in the graph. 
This allows for changing topologies and changing action spaces. 
Adding an additional host to a network means adding several more actions relating to the defense of that host.
For tabular methods, this would require fully retraining the agent; with our method, the action space is already a function of the graph size, so no retraining is required.


This work presents a generalizable framework for graph-based RL that allows for actions upon nodes and edges, with variably sized graphs. 
To the best of our knowledge, our approach is the first time relational inductive bias has been applied to ACD. 
We will show that our approach finds defense strategies that generalize better than prior work in the same field. 
The generality afforded by relational inductive bias means our agents can be deployed to new environments without retraining, saving potentially days of training time. 
In a simulated network environment, our agents score more than 4x higher than prior works and maintain that high score across environments with varied numbers of hosts--something the prior works are unable to do at all.
In more complex environments, we show that when we perturb the topology or introduce new adversaries, our models perform better than all prior works.
Finally, we show that our approach is also applicable to multi-agent reinforcement learning problems.
Ours is currently the highest performing non-heuristic policy in the CC4~\citep{cc4} environment.
The source code for the agents and experiments is available at \url{https://github.com/cybermonic/ACD-With-GraphRL}


    \section{Related Work}
\label{sec:rel}

\textbf{Relational Inductive Bias}: As in traditional machine learning, an overfit RL agent will optimize its policy to the random noise in the environment rather than its true distribution. 
This reflects a high degree of variance in the model. 
To counter overfitting, one must increase the model's bias~\citep{bias_variance_tradeoff}. 
Inductive bias helps models constrain their search space and reason with more generalized approaches~\citep{bias_in_learning}.
Inductive bias can be created implicitly via the choice of neural architecture, or explicitly by constraining the data the model receives. 
Motivated by the strong arguments of \citet{gnn_motivation} in favor of relational inductive bias, as well as the numerous positive results from empirical studies~\citep{gluing_blocks,globalnode,rl_tsp,graph_joints} we represent the networks we wish to defend as graphs.

\textbf{RL for Automated Cyber Defense}:
There exist many agents trained for cyber-tasks that focus on individual entities within networks, or abstract networks into individual units to defend~\citep{attack_attempts_viewable,rl_malware_creation}. 
When networks are represented as graphs, they are commonly abstracted into tabular representations for simplicity~\citep{booker2020model,cage1_hppo,ql_defenders}. 
Works that include graph representations often focus on finding vulnerabilities~\citep{attacks_and_defenses,rl_attack_graph_analysis} or automated penetration testing~\citep{crown_jewels,cyber_battle_sim_decoy_experiment}. 
While many graph-based ACD agents exist in the literature, they are often constrained in unrealistic ways. 
Prior works,~\citep{ddos_prevention_dqn,sdn_dqn,simple_experiment,graph_lstm_acd} all consider the graph structure of networks, and train DQNs to defend them. However, they all fail to consider what happens if nodes are added or removed, or even if the order in which nodes are indexed changes. 
More relevant to our work are the ACD approaches which highlight the importance of generalization to new topologies. 
One method to achieve this is aggregation. \citet{honeypot_dqn} create a generalizable agent by creating an action space that is independent of node count and order. 
Instead, when an action is selected, it is applied to the node that can utilize it most efficiently according to a value function. 
Similarly, \citet{ddos_very_compressed} aggregate the entirety of the network into a single entity to be defended; their observations are the network's state in aggregate across $k$ timesteps. 
However, it would be difficult to directly apply either of these approaches to the more general ACD environments we evaluate, as they are both specialized for narrow threat models. 
The most similar prior work to ours is \citet{doorman}. 
They model the network as a graph and train a DQN to select triplets of nodes to rewire for network hardening. 
Their approach is specific to this one task, however, and as we will show, it does not generalize well to more varied action spaces, but it is notable in that it is fully inductive.

\comment{
Another potential solution is multi-agent reinforcement learning (MARL); prior work~\cite{marl_ddos} assigns a new agent to monitor each important node in a network for. However, MARL is complex and introduces an unnecessary deficit of information for our use-case. 
Closer to our approach, Booker et al.~\cite{booker2020model} test the effect of changing node features, and their model's performance on 3 separate topologies--though the model is retrained for each new environment. The two approaches most similar, and most influential to our work, are ACD-G~\cite{acd_g} and the work of Wolk et al.~\cite{keeping_it_rl}. 
Both works test the robustness of their agents by evaluating them in environments with different node features and edges without retraining them. 
In particular, ACD-G generates 50 random graphs to evaluate pre-trained agents in. 
Wolk et al., because of the complexity of the CAGE-2 environment~\cite{cage2}, manually engineer new environments with re-indexed nodes, and edge perturbations to evaluate their pre-trained models. 
}

    \section{Background}
\label{sec:bg}

A graph is a set of discrete objects $\V$ called nodes, and a set of edges $\E = \{(u,v) \mid u,v \in \V\}$ denoting relationships between them. 
An attributed graph is a graph with features associated with each node, $f:\V \rightarrow \IR^d$.
We denote these features as $\mathbf{X}$. 
Finally, we define node embeddings as the output of a function $\Phi(\V, \E, \mathbf{X})$ that processes the graph to encode node features and topology into a single vector. 
This function can be transductive, which means it can only produce embeddings for nodes it has seen during training, or it may be inductive, which means it can produce embeddings for nodes it has never seen before. 
We will evaluate both kinds of embedding functions. 

A partially observable Markov decision process (POMDP) is defined as the 7-tuple
\begin{equation}
    \mathcal{M} = \langle\mathcal{S,A,O,R,T},\phi,p\rangle.
\end{equation}
Here, $\mathcal{S}$ is the set of states, $\mathcal{A}$ is the set of actions, and $p(s_0)$ is the distribution of possible starting states. 
$\mathcal{T}(s^\prime \mid s,a)$ is the possibly stochastic transition function that determines the next state given a state and action. 
$\mathcal{R}: \mathcal{S}\times\mathcal{A} \rightarrow \IR$ is a scalar reward function. 
The set $\mathcal{O}$ is the observation space, which is derived from the true state via the function $\phi: \mathcal{S}\rightarrow \mathcal{O}$. 
Any policy acting upon $\mathcal{M}$ will only observe the output of $\phi$.

\citet{zsg_survey} define context-based POMDPs (CMDPs) as the set of all possible POMDPs in a space conditioned by a parameter sampled from distribution $C$. 
This parameter changes the possible state space, transition probabilities, and rewards, but not the action space. 
If we divide the context space into disjoint sets, $\mathcal{M}|_{C_{\text{train}}}$ and $\mathcal{M}|_{C_{\text{test}}}$, the objective is to find policy $\pi$ which maximizes the reward over CMDPs parameterized from the test space conditioned only on experiences from the training space.


\comment{
\textbf{Problem statement}: We represent network defense as a two-player CMDP, where the context parameter is the network's connectivity, or topology. 
By doing so, we can define the context space as a set of graphs representing various network configurations. 
Agents may be trained in environments conditioned on a small subset of these graphs, and evaluated on the full set. 
We are interested in evaluating how well the skills a defensive agent learns in its training environment can be applied to the environments it is evaluated in.
This will be measured by evaluating the magnitude of the change in score between training and testing. 
}
    \section{Graph-based Reinforcement Learning }
\label{sec:method}
In this section, we will describe the architecture of our graph-analytic models. 
We will first describe how we represent observations as graphs and actions as graph edits.
Next, we describe the three models we evaluated: one transductive model, and two inductive models. 
Transductive models assume the size and node-ordering of a graph is fixed, while inductive models can adapt to unseen environments, with varying numbers of nodes. 
We used GCN as the GNN architecture, but this approach is agnostic to the specific choice of GNN.\footnote{Experiments showed low variance between different choices of GNN architectures. We provide an ablation study demonstrating this in Appendix~\ref{sec:appendix_ablation}.}

\subsection{State-Action Space}
\textbf{States}:
In all experiments, we use individual hosts as nodes, and forms of inter-host communication as edges. 
Other entities within a network can also be modeled as nodes in a graph, with their node type specified as one of their node features.
Unlike prior works which use features derived from graphs as their observations~\citep{acd_g,keeping_it_rl}, we use the attributed graph itself. 
This means agents' observations are represented by the 3-tuple $\mathcal{O} = \langle \mathbf{\V, \E, X} \rangle$, representing nodes, edges, and node features, respectively. 
Importantly, we do not assume the agent has full visibility into the network. 
The graph provided in the observation may have missing information about whether hosts are compromised, the possible paths an attacker could use to pivot through the network, etc., such that $\mathcal{O} \subseteq \mathcal{S}$. 

\textbf{Actions}:
Following the model of~\cite{globalnode}, we model each game as a system of objects (or nodes) that have relationships (or edges) with one another. 
Actions can be performed upon some subset of nodes to change the environment. 
Thus, for each actionable node $v_i\in \V_a \subseteq \V$, given an action space $\mathcal{A} = \{a_0, ..., a_n\}$, actions are represented as functions upon those nodes, $a_n(v_i)$. This makes the total action space for an environment $\mathcal{A} \otimes \V_a$. By representing actions as functions upon discrete objects, rather than a fixed array as is done by tabular methods~\citep{qlearning}, we are free to increase or decrease the action space without retraining by changing the size of set $\V_a$.

The results of actions produce graph edits. These edits may be in the form of node or edge additions or deletions, or changes in nodes' features. For example, the action $a_n(v_i)$ may create a node $v_j$ with an edge to $v_i$; it could change the feature vector $x_i$ associated with node $v_i$; or it could remove the node $v_i$ from the graph entirely. With this abstraction, we can further extend the action space to include edge-level actions as a function of $a_n(v_i, v_j)$, upon actionable edges $\E_a$. 
This work considers environments that have node-level and edge-level actions, $\mathcal{A} = \mathcal{A}_{\V} \cup \mathcal{A}_{\E}$.

\subsection{Agent Design}
To combine the information about the graph structure, and its features, we use a graph neural network (GNN)~\citep{gnn} to produce node embeddings. We employ a graph convolutional network (GCN)~\citep{gcn} for its expressiveness, and inductive abilities. The defining function of the GCN is
\begin{equation}
    \mathbf{H}^{(\ell+1)} = \sigma\big(\mathbf{D}^{\frac{1}{2}}\mathbf{AD}^{\frac{1}{2}}\mathbf{H}^{(\ell)}\mathbf{W}^{(\ell)} + \mathbf{b}^{(\ell)}\big)
\end{equation}
where $\mathbf{A}$ is the adjacency matrix of the graph plus $\mathbf{I}$. $\mathbf{D}$ is the degree of each node, and $\mathbf{W}^{(\ell)}$ and $\mathbf{b}^{(\ell)}$ are trainable parameters. For each node in the graph, this function averages the features of its neighbors, then passes the output through a single fully connected nonlinear neural network layer. Using $k$ GCN layers will encode information about each node's $k$-hop neighborhood. In practice, we use the more efficient message-passing paradigm implemented by PyTorch Geometric~\citep{torch_geo}. 

The agent learns via Proximal Policy Optimization (PPO)~\citep{ppo}. 
Both the actor- and critic-networks create node embeddings $\mathbf{Z}$ using 2-layer GCNs. 
The embeddings are then passed through additional fully connected layers to produce the probability distribution function, or the state value estimate for the actor and critic, respectively. 
We evaluate three methods to convert from node embeddings to action probabilities. 
Using the terminology of graph representation learning, we refer to these approaches as either inductive or transductive~\citep{sage}.

\begin{figure}[t]
    \centering
    \begin{minipage}{0.5\textwidth}
    \centering
    \includegraphics[width=0.9\linewidth]{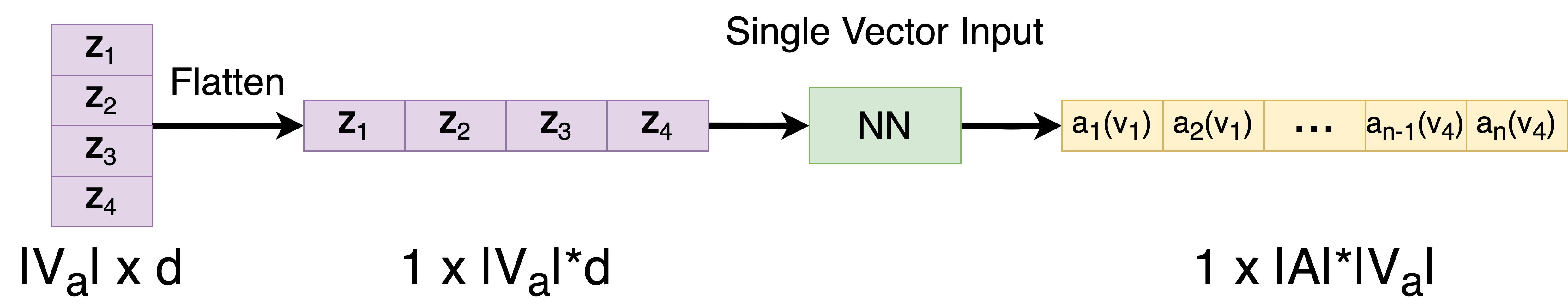}
    \caption{Transductive Actor}
    \label{fig:transductive}
    \end{minipage}%
    \begin{minipage}{0.5\textwidth}
    \includegraphics[width=0.9\linewidth]{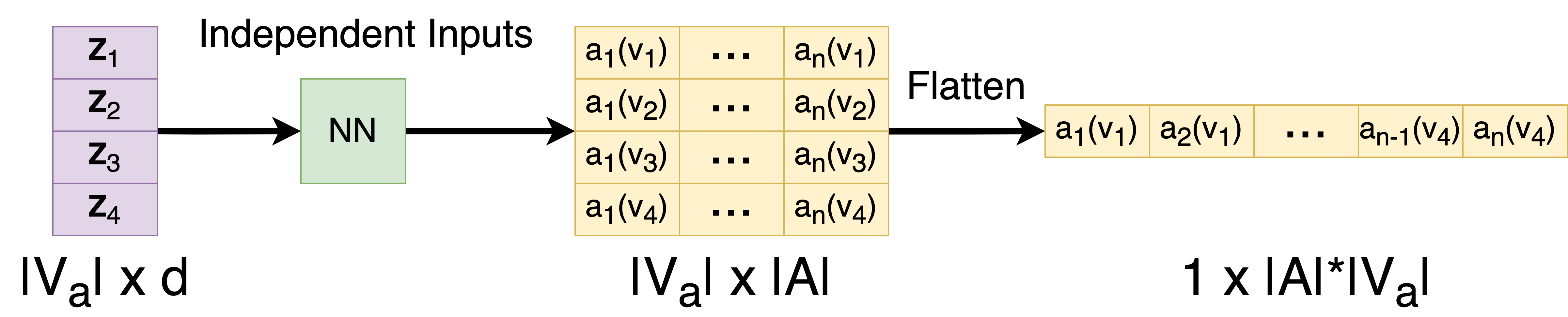}
    \caption{Inductive Actor}
    \label{fig:inductive}
    \end{minipage}
\end{figure}

Transductive models operate as prior tabular RL approaches do. 
As shown in Figure~\ref{fig:transductive}, they concatenate the node embeddings into a single, fixed-sized vector, and pass it through a traditional neural network to produce a single vector of size $|\V_a|\cdot|\mathcal{A}|$. 
This approach is not generalizable to new topologies. 
However, as transductive models tend to outperform inductive ones~\citep{transductive_v_inductive}, it serves as a basis to demonstrate the maximum potential of relational bias in this domain. 

For inductive models, the embeddings for each node must be processed in an order- and length-invariant way. 
The simplest way to do this is illustrated in Figure~\ref{fig:inductive}. 
The model passes each node embedding independently through a fully connected layer, and the output is then flattened. 
The actor-network uses the softmax of the flattened output to directly calculate action probabilities (e.g., $v_{i,j}$ represents the log-odds of taking action $j$ on node $i$). 
If more actionable nodes are added to the graph, the inductive GNN will output a $|\V \cup \V^\prime| \times |\mathcal{A}|$ dimensional matrix, which can be interpreted in the same way without retraining, thus allowing for full inductivity. 
The critic-network, which attempts to evaluate the value of the current global state, projects each row of $\mathbf{Z}_{\V_a}$ into a single dimension, then pools the batch into a single value. 
This method, which we refer to as the \textit{naive inductive model} is simplistic, but we will show it is also powerful. 
However, it has a major drawback: node embeddings do not have the context of other nodes' states if they are greater than $k$ hops away. 

To address this, we adopt a modified form of the attention-based node pooling proposed by~\citet{globalnode}. 
This modification occurs in the node embedding step of the model. 
In each layer of the GCN, the embeddings for all nodes $\mathbf{Z}$ are calculated. 
Then, the embeddings of actionable nodes, $\mathbf{Z}_{\V_a}$, are extracted and concatenated with a global graph state vector $\mathbf{g}$. 
We then calculate the updated global state vector of the graph as 
\begin{equation}
    \mathbf{g}' = \mathbf{g} + \phi_g\Big(\text{POOL}_{i\in \V_a} \{ \phi_v(\mathbf{g},\mathbf{z}_i) \cdot \phi_a(\mathbf{g},\mathbf{z}_i) \} \Big)
\end{equation}
where $\phi_v$, and $\phi_g$ are fully connected networks, and $\phi_a$ is a fully connected network with softmax activation. 
POOL represents any pooling function that is order-invariant. 
The original work suggests sum-pooling, but this can cause issues when testing in environments that have many more nodes than the training environment. 
In this work, we use mean- or max-pooling to address this problem. 

The actor-network concatenates the final $\mathbf{g}$ vector to each of the final node embeddings, then uses the final vectors in the inductive method we previously described. 
This ensures that each node embedding contains information about all other nodes in the network, which allows them to weigh the importance of taking an action upon themselves, vs. an action somewhere else in the network. 
The critic network, as it is calculating the global value of the network's state, uses the $\mathbf{g}$ vector directly as input. 
We refer to this method as the \textit{attention inductive model}.

Inductive models also support \textit{edge-level actions}. 
We formulate the probability of taking an action on an edge as the output of an additional function that takes the source and destination node embeddings as input. 
In this work, to calculate the probability of taking an edge action, we calculate
$f(\mathbf{Z}_{\text{src}} \odot \mathbf{Z}_{\text{dst}})$ where $f(\cdot)$ is a fully connected layer with $|\mathcal{A}_{\E}|$-dimensional output, $\odot$ represents the Hadamard product, and $\mathbf{Z}$ is the set of node embeddings such that $\langle \text{src},\text{dst} \rangle \in \E_a$. 
The output of $f(\cdot)$ is then concatenated to the flattened probability vector of node-level actions. 
    \section{Experiments}
\label{sec:results}

We evaluate our agent in three environments of increasing complexity.
In each environment, our agent (the blue agent) defends the network from an attacker (the red agent). 
The threat model between each environment varies, but broadly, we assume that the system can be modeled as a graph of hosts whose security states transition over time between “safe” and “compromised.” These transitions are governed by both attacker and defender actions. The specific transition dynamics vary by environment but are always represented within the environment’s MDP.
We assume that the attacker begins with root access to a single host and can take actions that increase the likelihood of compromising additional hosts. The attacker operates under partial observability of the network, must scan to discover reachable hosts, and lacks prior knowledge of the topology. 
The defender also has partial observability, though the scope and quality of information vary by environment.
In each timestep, both attacker and defender select actions simultaneously. 
The attacker's objective is to maximize the number of compromised hosts over a finite time horizon, while the defender aims to minimize this quantity. 

Table~\ref{tab:envs} summarizes important details of each environment.
For the first experiment, we analyze the Yawning Titan environment~\citep{acd_g}, one of the first works to frame RL-based network defense as a graph problem.\footnote{We select this work rather than \citet{simple_experiment} because it has an additional focus on domain generalization.}
Next, we evaluate on the CAGE Challenge 2 (CC2) environment~\citep{cage2}, which has a more complex action space and a more fine-grained reward function. 
Finally, we test our agent in the very complex CC4 environment~\citep{cc4}. 
In addition to being the largest network we test our agent on, it is also a multi-agent RL environment. Due to the page limit, please refer to the appendix for additional details on training configurations, ablation studies, and the details of the testing environments.

\begin{table}[h]\centering
\caption{Evaluation Environments}\label{tab:envs}
\scriptsize
\begin{tabular}{lccccc}\toprule
&Network Size &Per-node actions &Per-edge actions &Blue Agents &Red Agents\\\midrule
Yawning Titan &10-100 &3 &0 &1 &1\\
CC2 &13 &10 &0 &1 &1\\
CC4 &32-128 &4 &2 &5 &1-5\\
\bottomrule
\end{tabular}
\end{table}

\subsection{Network Simulation with Yawning Titan Environment}
\label{subsec:yt}

This environment is a two-player CMDP where a heuristic red agent attempts to spread through a computer network, and our model attempts to defend it. 
The computer network is represented as a random Erd\H{o}s-R\'enyi graph~\citep{erdos_renyi}.
Each node represents a computer, and edges represent their ability to communicate. 
Details of the experimental setup and the state-action space of this environment are provided in the Appendix. 

\comment{
\begin{figure}[t]
    \centering
        \includegraphics[width=\linewidth]{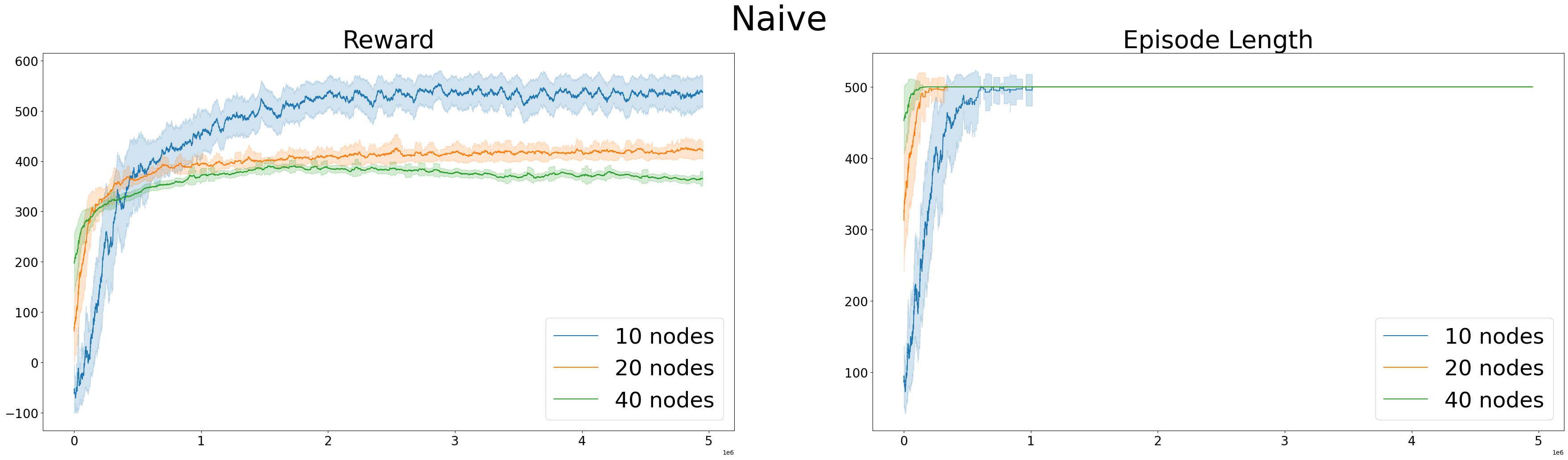}
        \includegraphics[width=\linewidth]{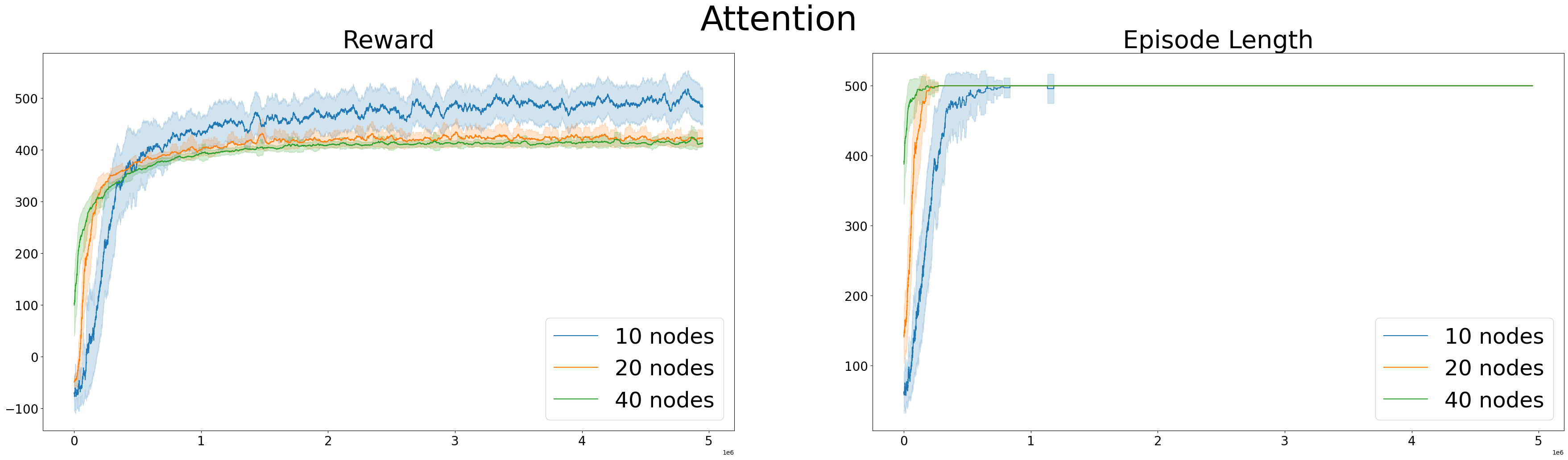}
        \caption{Average reward over time during training for the naive and attention-based models. Shaded areas around the lines represent standard deviation.}
        \label{fig:yt_training}
\end{figure}
}

We use the same training configuration as the original work~\citep{acd_g}. 
We train models for 5 million steps on a single random graph, then sample 50 new graphs and evaluate the models in these new environments without retraining for 10 episodes each.
We compare our \textit{Transductive}, \textit{Naive}, and \textit{Attention Inductive} models to the two approaches from the prior work: \textit{Standard Observations}  and \textit{Graph Observations}. 
We also evaluate the architecture proposed by \citet{doorman}: a struc2vec-based GNN~\citep{struc2vec} that uses sum aggregation to produce a global graph vector that is used in a similar way to our self-attention model. 
Their original approach was only designed for environments with a single action, but we modified the output to produce likelihoods for $|\mathcal{A}|$ actions per-node. 
Their architecture is similar to ours, but their approach uses a DQN rather than PPO. 
It is included here to compare the utility of both RL strategies for this problem. 
Table~\ref{tab:yt_results} reports the average reward of all 500 episodes, normalized such that 1.0 is a theoretically perfect score if no hosts were compromised at all. 

\begin{table}[htbp]
\centering
\caption{Mean score on random environments (higher is better)}\label{tab:yt_results}
\scriptsize
\begin{tabular}{llll}\toprule
&\multicolumn{1}{c}{$|\V| = $ 10} &\multicolumn{1}{c}{$|\V| =$ 20} &\multicolumn{1}{c}{$|\V| = $ 40} \\\midrule
Standard Observations &0.2511 $\pm$ 0.017 &0.2310 $\pm$ 0.013 &0.1683 $\pm$ 0.010 \\
Graph Observations &0.2090 $\pm$ 0.005 &0.3037 $\pm$ 0.013 &0.2047 $\pm$ 0.013 \\
\citet{doorman} &0.7379 $\pm$ 0.011 &-0.0385 $\pm$ 0.011 &-0.0300 $\pm$ 0.007 \\
\midrule 
Transductive &0.4246 $\pm$ 0.011 &-0.1327 $\pm$ 0.000 &-0.0966 $\pm$ 0.000 \\
Naive Inductive &0.2681 $\pm$ 0.036 &0.5439 $\pm$ 0.012 &0.9308 $\pm$ 0.004 \\
Attention Inductive &\textbf{0.8167} $\pm$ 0.009 &\textbf{0.5634} $\pm$ 0.012 &\textbf{0.9955} $\pm$ 0.002 \\
\bottomrule
\end{tabular}
\end{table}
 
Unsurprisingly, the transductive model is unable to generalize to new environments. 
It is likely that it overfit to the node ordering it observed during training.
Both the naive and attention-based inductive models more than double the tabular methods' scores. 
When compared to the other GNN-based technique, as environments grew more complex, the model became less able to generalize to new graphs. 
Despite achieving scores comprable to our models during training, when evaluated on the new graphs, the \citet{doorman} models failed to generalize. 
This may be explained by PPO's greater exploration ability, allowing it to find more optimal policies in more complex environments compared to DQN~\citet{dqn_ppo_a2c}.
It could also be that the prior work's choice of sum for the readout function in their model did not generalize to graphs with varied densities, leading to oversmoothing. 
We also observe that when $|V|=40$, the attention-based model appears to have found a nearly perfect strategy. 
We hypothesize that because it was trained in a more complex environment, the observations it received during training were more varied, and forced the model to find a more general policy, while models in smaller environments became more overfit. 

\begin{figure}[h]
    \centering
    \includegraphics[width=0.5\linewidth]{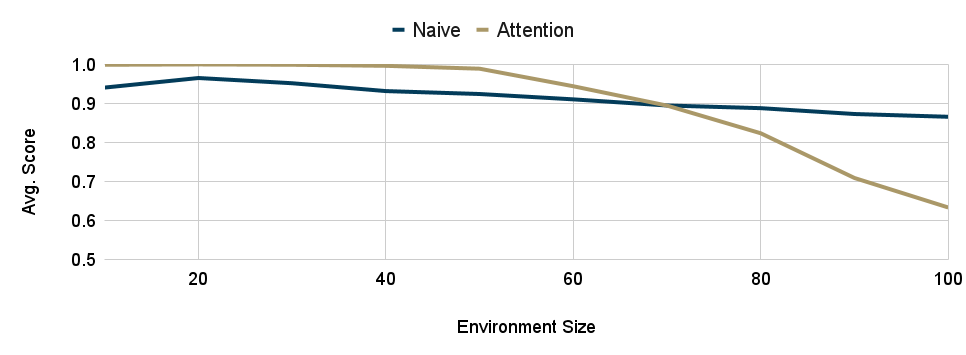}
    \caption{Scores attained by the $|\V|=40$ GNN models on graphs of different sizes.}
    \label{fig:yt_new_sizes}
\end{figure}

To further demonstrate the generalization of the policies our models learn, we evaluate the high-scoring $|\V|=40$ models in environments with different numbers of nodes than they were trained in. 
Unlike the prior work, which cannot generalize to different environment sizes, our approach is fully inductive.
As before, we evaluated the model without retraining 10 times on 50 randomly sampled graphs for each environment size. 
Figure~\ref{fig:yt_new_sizes} plots the scores of the naive and attention-based models in these different environments.

Both models, even in the largest environment we tested, still more than double the scores of the Euclidean models in the simplest environments. 
The attention-based model achieves higher scores overall: its strategy remains near-perfect in environments where $|\V| \le 50$. 
However, it is more sensitive to the growing complexity of larger environments than the naive model is. 
While the naive model has slightly lower scores than the attention-based model, it appears that its simpler design allowed it to remain generalizable for much more complex environments. 
Its score also decays as complexity increases but at a much lower rate than the attention-based model. 
These results are likely a result of the No Free Lunch theorem~\citep{nfl}, which states that any increase in a model's performance in one problem domain must be accounted for by a deficit elsewhere. 
These results seem to indicate that the cost of good model generality may be lower performance in easier regions of the environment-space, but consistent performance across more of the space. 

\subsection{CAGE-2: A Simulated Enterprise Network}
\label{sec:cage}

\comment{
\begin{figure}[b]
    \centering
    \includegraphics[width=\linewidth]{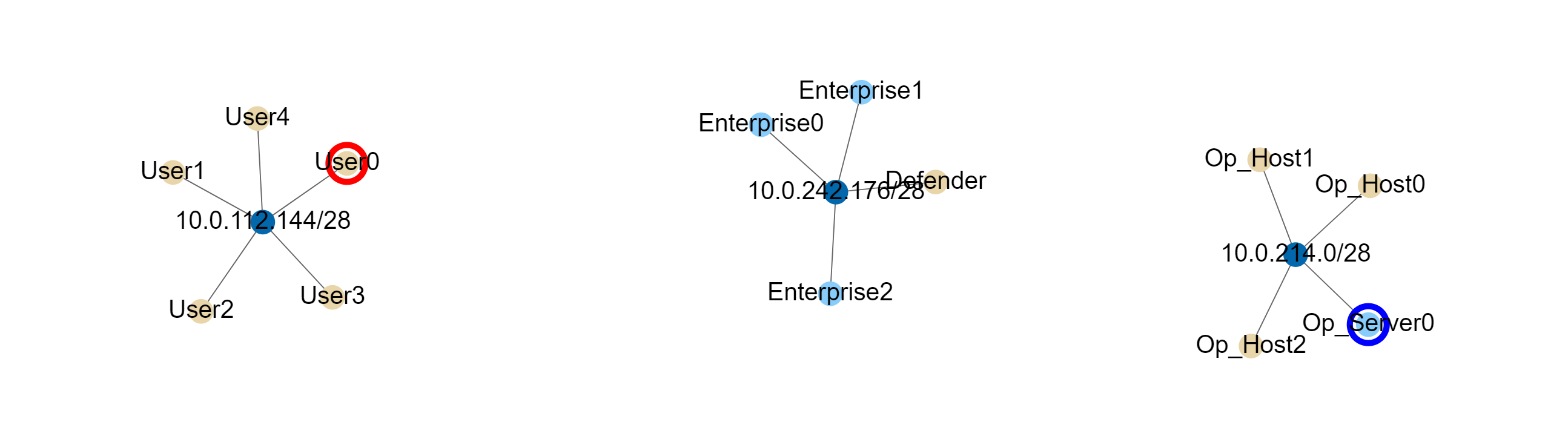}
    \caption{The three default subnets in the CAGE-2 topology. Hosts are light blue, and subnets are dark blue. The host circled in red is compromised by default. The server circled in blue is OpServer0, which we do not want the red agent to capture.}
    \label{fig:cage_graph}
\end{figure}
}

The CAGE Challenge-2 (CC2) was a reinforcement learning contest that aimed to ``support the development of AI tactics, techniques, and procedures...for autonomous cyber operations"~\citep{cage2}. 
In this scenario, the red agent is trying to reach and compromise a specific, important host called OpServer0. 
CC2 provides two heuristic-based red agents: B-Line, and Meander. 
The B-Line agent knows the fastest route from its starting machine to its target, OpServer0, and will take the same path of lateral movements through the network with little variation. 
The Meander agent is slower, performing a breadth-first search through each subnet before moving to the next subnet. 
Specifics about the state-action space and reward function are provided in the Appendix.

\textbf{Training:}
All GNN models are trained using the same hyperparameters and settings. 
For each episode, a red agent is selected randomly with even probability. 
We then simulate 100 timesteps. 
After 100 episodes, the agent's weights are updated according to the PPO algorithm~\citep{ppo}. 
The agent only uses 4 epochs per update and otherwise uses the same hyperparameters as StableBaselines3~\citep{stable_baselines3}.
Both the actor and critic networks use two-layer GCNs with 256- and 64-dimensional layers.\footnote{An ablation study on this hyperparameter is provided in the Appendix.}
The attention-pooling models use a 256-dimensional global vector with mean-pooling. 
All models were trained for 100k, 100-step episodes. 

\begin{table}[h]\centering
\caption{Scores on the CC2 Environment (smaller is better)}\label{tab:scores}
\scriptsize
\begin{tabular}{lrrrrrrrr}\toprule
& &\multicolumn{2}{c}{30 Steps} &\multicolumn{2}{c}{50 Steps} &\multicolumn{2}{c}{100 Steps} \\\cmidrule{3-8}
&Total &B-Line &Meander &B-Line &Meander &B-Line &Meander\\\midrule
Cardiff &\textbf{-54.57} $\pm$ 0.43 &\textbf{-3.47} &\textbf{-5.64} &-6.41 &\textbf{-8.69} &-13.76 &-16.60 \\
Keeping it RL &-56.90 $\pm$ 0.59 &-3.48 &-6.47 &\textbf{-5.85} &-10.33 &\textbf{-11.39} &-19.38 \\\midrule
Transductive &-57.30 $\pm$ 0.65 &-3.58 &-6.25 &-6.59  &-9.86 &-14.09 &-16.94 \\
Naive Inductive &-62.08 $\pm$ 0.63 &-4.16 &-7.20 &-7.21 &-10.90 &-15.15 &-17.45 \\
Attention Inductive &-60.14 $\pm$ 0.66 &-4.39 &-6.99 &-7.68 &-9.80 &-14.80 &\textbf{-16.48} \\
\bottomrule
\end{tabular}
\end{table}

\textbf{Default Game:}
Each agent was evaluated on 100 episodes with lengths $\{30, 50, 100\}$ against one of the heuristic red agents.
The average score of these episodes is reported in Table~\ref{tab:scores}.
We compare our models to the top two performers from the original competition: \textit{Cardiff}~\citep{cardiff}, an HPPO approach which trains expert agents against the B-line and Meander red agents, and heuristically decides which model to use, and \textit{Keeping it RL}~\citep{keeping_it_rl} an ensemble-of-ensembles of PPO models trained against different red agents. 
These models outperform ours in the competition evaluation, but not by a wide margin.

As expected, because the evaluation environment is the same as the training environment, the transductive model achieves the best average score of our models. 
However, the attention-based model scores almost as high in several instances, and even better in the 100-step game against Meander.
Additionally, the difference in performance between the inductive and transductive models is minor.

\begin{figure}[t]
    \centering
    \begin{minipage}{0.5\textwidth}
        \centering
        \includegraphics[width=0.8\linewidth]{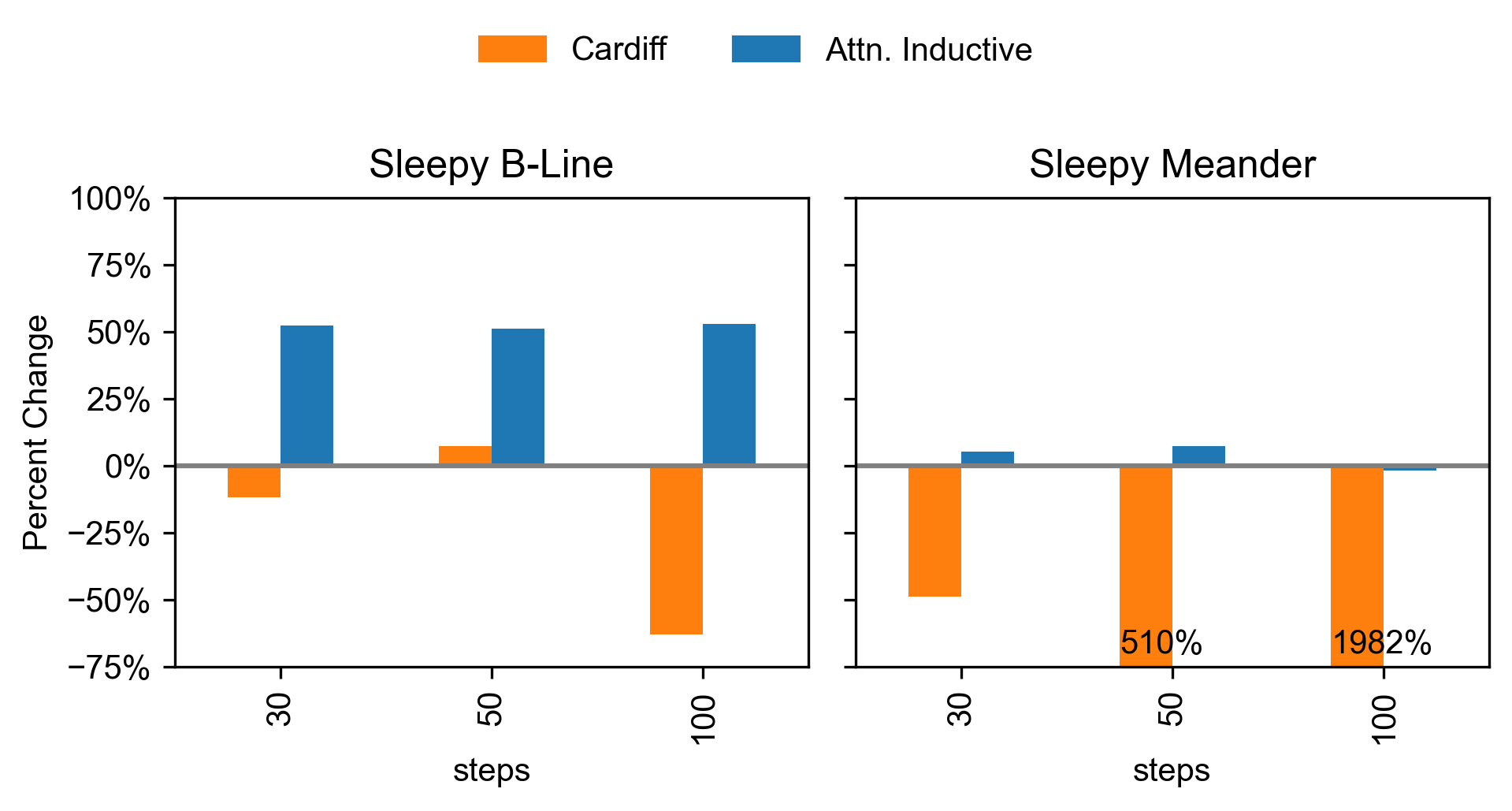}
        \caption{Evaluation against unseen red agents}
        \label{fig:sensitivity}
    \end{minipage}%
    \hfill%
    \begin{minipage}{0.5\textwidth}
        \centering
        \vspace{0.7cm}
        \includegraphics[width=\linewidth]{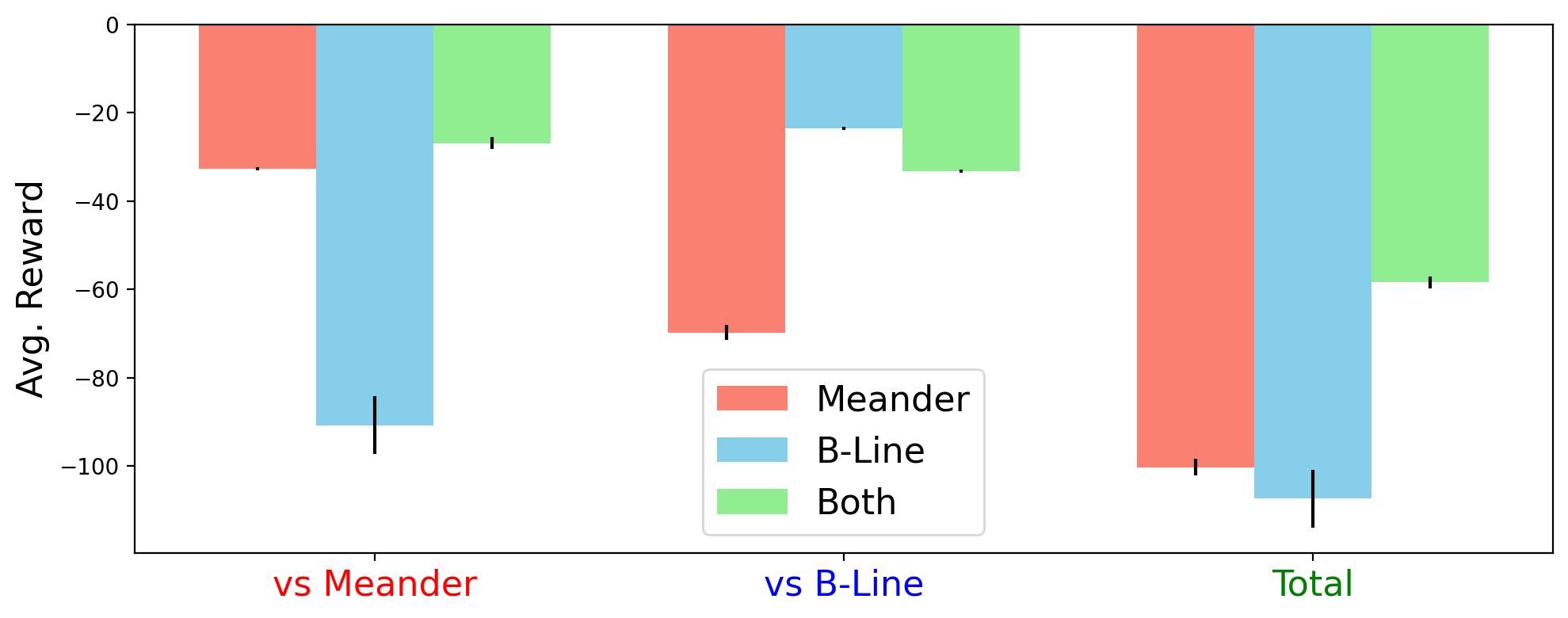}
        \caption{Evaluation of models trained against a single red agent tested against both red agents}
        \label{fig:specialists}
    \end{minipage}
\end{figure}

\textbf{Adversary Generalization:} While the top models are very good at this specific task, they are also very brittle and overfit to the environment. 
In a realistic scenario, we expect the adversary to act in unpredictable ways. 
We create two new red agents: Sleepy-Meander, and Sleepy-B-Line. 
These are slower red agents; the only difference in their strategy is that at every turn they have a 50\% chance of selecting no-op instead of the move the non-sleepy agent would have taken. 
Compare this to training against automated attacker agents, and the defense agent encountering a slower, human attacker for the first time. 
We evaluate our transductive model and the CardiffUni model against these new agents with no retraining.
The results in Figure~\ref{fig:sensitivity} show that this minor, realistic perturbation caused the Cardiff agent to experience a 1,982\% decrease in its score in the worst case. 

\comment{
Because the top agents are hierarchical, they learn to identify which of the two red agents it was trained on is currently attacking. 
This works exceedingly well against agents these models have seen before, but when a new red agent appears, they are unprepared. 
}
In comparison, utilizing relational inductive bias allows our agents to learn more generalizable policies, which is evident in how they can take advantage of the weaknesses of new attackers. 
For example, because the Sleepy B-Line agent is 50\% slower, our model achieves a 50\% better score. 
Against the slower meander agent, the Cardiff HPPO agent selects the expert policy for B-Line and suffers immensely. 
On the other hand, our agent applies its more universal policy that it learned from graph analysis and scores slightly better, or about equal to before.
These results show that relational inductive bias allows for generalization to new attackers; next, we will evaluate our agents' ability to generalize to new environments. 

In another experiment, we trained our blue agents only against a single red agent before testing them against both red agents.
The results of this study are shown in Figure~\ref{fig:specialists}.
We found that against unseen agents, our models perform comparably to the HPPO agent evaluated by~\citet{keeping_it_rl}. 
Interestingly, the agent trained against only Meander scores slightly lower against the Meander agent, compared to the baseline agent that was trained against both red agents. 
These results suggest that the more general policy the default agent learned to defend against both red agents is stronger than the locally optimal policy the Meander-specialist discovered to defend against a single agent.

\textbf{Domain Generalization:}
\label{subsec:generalization}
\comment{Another realistic concern is the agent's ability to defend a differently sized network than it was trained on. 
Suppose a company hires $k$ new members; this will add at least $k$ new hosts to the network. 
It would be inefficient and expensive to retrain their ACD agent every time a hiring or firing occurs, or even if a new device joins the network.
Luckily, inductive models are adaptable to graphs of varied cardinalities. 
Adding and deleting nodes will not affect them in the same way it would affect a tabular method, which always expects the same number of hosts as input and always outputs the same $|\V_a||\mathcal{A}|$-dimensional action vector.}
In addition to new adversaries, it is important to evaluate how agents behave when faced with new environments. 
This concern about generalizability was shared by \citet{keeping_it_rl}, so they devised 3 new scenarios to evaluate their models in new environments. 
Scenarios 3 and 4 shuffle the order that hosts are indexed in two of the subnets; Scenario 5 adds paths from each machine in one subnet to two hosts in another subnet. 
These changes are made to the configuration file that generates the environment and are not explicitly communicated to the agents. 
For all generalization experiments, we evaluate agents with the parameters learned from the default CC2 environment without retraining.

\begin{figure}[t]
    \centering
    \includegraphics[width=0.5\linewidth]{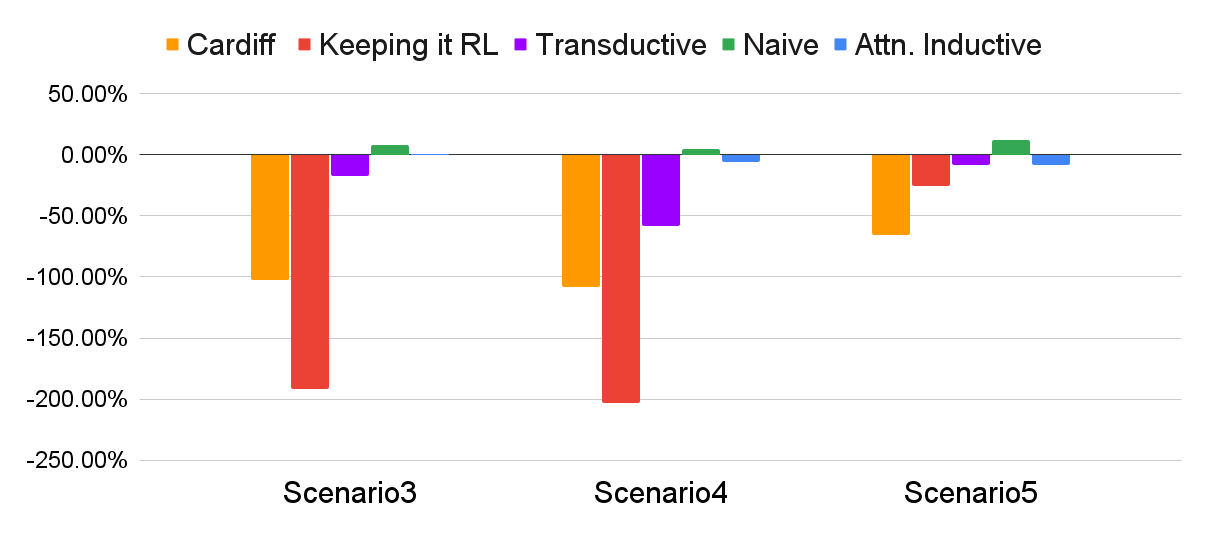}
    \caption{Change in average score under new scenarios proposed by ``Keeping it RL"~\cite{keeping_it_rl}}
    \label{fig:wolk_scenarios}
\end{figure}

Figure~\ref{fig:wolk_scenarios} shows how different agents respond in these new scenarios. 
We note that even our transductive model out-performs the prior works, highlighting the utility of relational inductive bias.
We further observe that, as the theoretical backing predicts, the inductive models are unaffected by the scenarios that change the ordering of nodes. 

The scenarios proposed by~\citet{keeping_it_rl} are a good starting point, but we feel that their changes do not go far enough. 
In addition to these scenarios, we also evaluate what happens when every index in the graph is perturbed.
We observed that both inductive models' scores changed by $<2\%$ for the index perturbation experiments, which was well within the standard error we observed in the previous experiments. 
However, the transductive model's average score dropped to -1,950.72.\footnote{For reference, selecting actions randomly will produce a score of 1999.41.}
From this, we conclude that the transductive model is also overfit. 

\begin{figure}[t]
    \centering
    \includegraphics[width=0.8\linewidth]{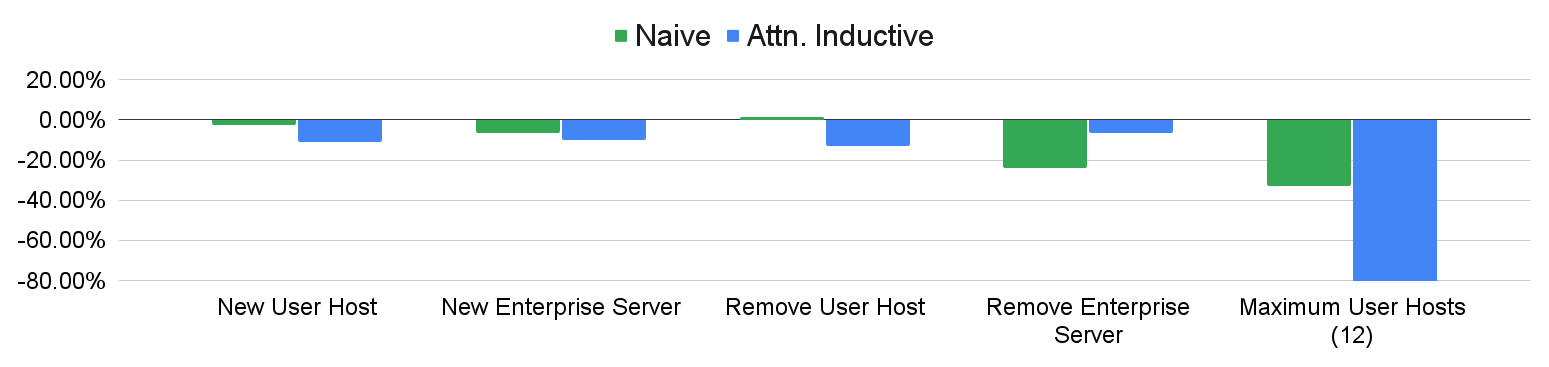}
    \caption{Change in inductive agents' scores in new environments with different numbers of hosts.}
    \label{fig:new_scenarios}
\end{figure}

Finally, we evaluate several new scenarios. 
These scenarios each involve a different number of hosts than were present in the training graph, so the tabular methods that we evaluated previously and the transductive model cannot run without retraining. 
Figure~\ref{fig:new_scenarios} shows how changing the environment affects the inductive agents' abilities to defend the network.
In the first four scenarios, we simply add or delete a node from a subnet. 
These minor perturbations in the network affect our agents very little. 
In both instances where a host is added, the agents' scores decrease slightly, but this is expected, as the attack surface increases with every additional host. 
Removing an enterprise server negatively affects the naive model, while having very little effect on the self-attention model. 
The naive model has only a local view of each node, so it does not know when to prioritize other subnets. 
As a result, it rarely attempts to defend the user subnet, opting instead to defend the Enterprise subnetwork. 
This inability to shift its attention likely explains the score decrease.

In the last scenario, where the user subnetwork is filled, both agents have difficulty defending the network, as the attack surface is nearly doubled: 8 new users are added.
Like the Yawning Titan experiments, the increase in the size of the environment affects the self-attention network more than the inductive network, which highlights the difficulty and importance of balancing good scores in simpler areas of the problem space with generality across regions with greater complexity.

\subsection{CAGE-4: Multi-Agent Reinforcement Learning}
The CAGE Challenge-4 (CC4)~\citep{cc4} extends CC2 into a Multi-Agent Reinforcement Learning (MARL) task. 
Now, multiple agents each defend one of five segmented networks, with actions and objectives similar to CC2. 
One key difference for this environment is its randomly initialized topology. 
During each episode, subnets are generated with 1-6 servers and 3-10 user hosts. 
Because networks are variably sized, prior transductive and tabular-based solutions will not work. 

The action space for this environment is similar to CC2. 
Some of the node-level actions from CC2 are merged into a single action, and there are two new edge-level actions: AllowTraffic and BlockTraffic, which create or delete edges between subnet nodes.
Specifics about the state-action space and reward function are in the supplemental material. 

\begin{figure}
    \centering
    \includegraphics[width=0.9\linewidth]{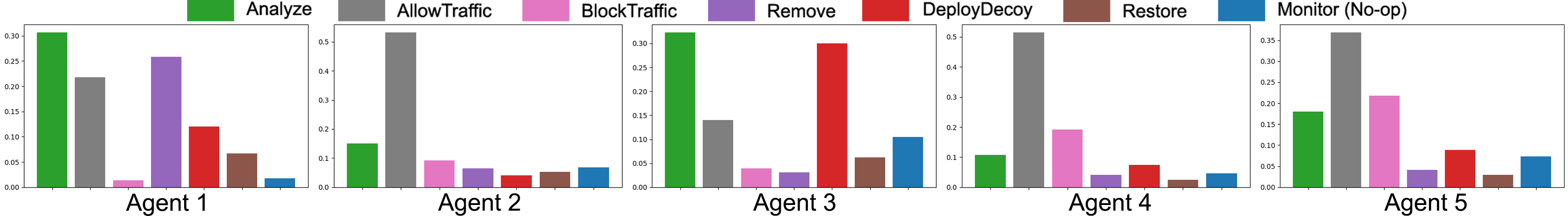}
    \caption{Action distributions of five CC4 agents. Each was trained independently, but interestingly, agents defending similar networks, e.g., Agents 2, 4, and 5 found similar strategies. Agent 1 defended the largest subnet, where it learned to analyze more frequently for better information about the large network it defended, while the latter four focused on enforcing firewall rules.}
    \label{fig:cc4_action_distro}
\end{figure}

We train five independent instances of our self-attention inductive model in this environment using the same configuration as in CC2 to measure our approach's applicability to MARL tasks. 
Each agent independently learns a unique policy for the subnet it defends. 
We illustrate the action distribution used by each agent in Figure~\ref{fig:cc4_action_distro}. 

Our approach was the highest-performing non-heuristic agent submitted to the CC4 competition and finished in fifth place overall~\citep{cage4_paper}. 
This result highlights the wide applicability of graph-based RL in fields beyond simple two-player games. 
Our agent's ability to adapt to variable network sizes and topologies allowed us to apply it to this new challenge with only minimal changes to accommodate the new action space. 
In a metareview of the competition, the competition runners identified the inability to adapt to the random initialization of the environment as the main detriment for other MARL approaches that were submitted~\citep{cage4_paper}.
Representing actions as functions upon nodes and edges allowed our approach to naturally adapt to this challenge. 
These  results show great promise for future research into multi-agent reinforcement learning with relational inductive bias. 

\comment{
\begin{table*}[!htp]\centering
\caption{Change in Score After Topological Perturbations}\label{tab: }
\scriptsize
\begin{tabular}{lrrrrrrrr}\toprule
& &\multicolumn{2}{c}{30 Steps} &\multicolumn{2}{c}{50 Steps} &\multicolumn{2}{c}{100 Steps} \\\cmidrule{3-8}
&Total &B-Line &Meander &B-Line &Meander &B-Line &Meander \\\midrule
$+1$ User Host &-2.737\% &7.005\% &-6.234\% &0.826\% &-6.017\% &-2.973\% &-2.833\% \\
$+1$ Enterprise Server &-6.322\% &0.475\% &-1.236\% &-16.59\% &-1.847\% &-9.421\% &-6.032\% \\
$-1$ User Host &1.315\% &-3.491\% &13.56\% &-7.485\% &10.28\% &-11.81\% &6.331\% \\
$-1$ Enterprise Server &-24.11\% &-21.68\% &-8.249\% &-34.56\% &-11.36\% &-35.86\% &-25.12\% \\
Max User Hosts &-32.75\% &6.151\% &15.28\% &-0.578\% &-57.37\% &-0.482\% &-87.04\% \\
Min User Hosts &3.762\% &-20.19\% &27.48\% &-18.44\% &26.13\% &-17.55\% &12.58\% \\
\bottomrule
\end{tabular}
\end{table*}
}

    \section{Discussion}
\label{sec:discussion}
\textbf{Limitations}. In this work we evaluate our approach on relatively small graphs. 
In all experiments, the bottleneck for throughput is the environment, rather than the model. 
However, with larger networks, more optimizations may be required to process observations at a reasonable speed. 
Another limitation is the leap from sim-to-real. 
Further engineering is required to address how the simplistic actions that agents take would be translated into real-world cybersecurity rules. 
Additionally, further analysis needs to be done to determine the utility of ACD agents, and how they would impact the humans using the networks they defend. 

\textbf{Future Work}. We only consider actions that manifest as node-level edits, with the exception of some simple edge-level actions in CC4. 
Future work on efficient representation of edge- and global-level actions within a graph-based RL framework is a promising next step. 
Additionally, future work could include new analysis on how to better implement intra-agent communication in MARL, and studies into performance improvements. 

\section{Conclusion}
\label{sec:conc}
As automated cyber defense becomes a more plausible option for businesses and governments, the high cost of retraining agents is an important constraint to consider. 
We have demonstrated that our proposed agent can generalize and defend never-before-seen networks from attackers with novel behavior. 
Because the agent understands the network as a graph and views its actions as graph edits, its policy is less sensitive to environmental perturbation. 
We show that our graph-based approach outperforms top RL approaches in many environments by a wide margin in the face of slight and major environmental perturbations. 
Our results show that relational inductive bias is a powerful tool for improving agents' generalizability, and a step toward ACD in real-world systems.  

    \bibliographystyle{iclr2025_conference}
    \bibliography{references}

    \appendix
    \newpage

\section{Training Configuration}
\label{sec:appendix_hyperparams}
To better facilitate replication of our work, in Table~\ref{tab:hyperparams}, we provide the hyperparameters used for each model. 
Due to the length of time required to train each model, we did not run ablation every hyperparameter, and instead opted for default settings in most cases. 
In the Yawning Titan environment, we selected 5M steps as the length of training time for fair comparison with the original paper. 
We selected the lower hidden and embedding dimension in Yawning Titan after initial tests with parameters identical to the CC2 environment performed poorly. 
In both environments, we tested both max and mean pooling and selected the model that performed best. 

\begin{table}[!htp]\centering
\caption{Model Hyperparameters}\label{tab:hyperparams}
\scriptsize
\begin{tabular}{lrrr}\toprule
&YT &CC2/4 \\\midrule
Hidden Dimension &64 &256 \\
Embedding Dimension &32 &64 \\
Pooling Function &Max &Mean \\
Actor LR &3e-4 &3e-4 \\
Critic LR &1e-3 &1e-3 \\
Episode Length &600 Steps &100 Steps\\ 
Episodes per Update (N) &10 &100\\
Training Time &5M Steps &1M Episodes \\
\bottomrule
\end{tabular}
\end{table}

All agents were trained on a server with an Intel Xeon Gold 6338 CPU using a maximum of 100 cores to simulate N episodes in parallel. Using a GPU would accelerate training slightly, but the main performance bottleneck was the environment simulators, which were CPU-only. 

\section{Ablation Studies}
\label{sec:appendix_ablation}
In this section, we present ablation studies on the number of parameters used by the models, and the choice of GNN. 
For all ablation experiments, due to the length of time required to train models for 1M episodes, we constrain training time to 100k episodes. 
Otherwise, all parameters are fixed, and the same as they were for the original CC2 experiments other than the hyperparameter being ablated. 

\textbf{Hidden Dimensions.}
In these experiments, in addition to changing the hidden dimension, we also kept the ratio of hidden to embedding dimensions the same (4 to 1), such that the agent with a 32-dimensional hidden dimension has an 8-dimensional embedding, and the agent with 1028 hidden dimensions has a 256-dimensional embedding. 
Otherwise, all other hyperparameters are the same as before. 
We observe that as more parameters are added to the models, their average score does increase somewhat. 
However, adding additional dimensions comes at a cost: as models grow larger, the time it takes to perform forward and backward passes has a steep increase.
Our choice in selecting 256 as the hidden dimension was to strike a good balance between score and efficiency. 

\begin{figure}[h]
    \centering
    \includegraphics[width=0.75\linewidth]{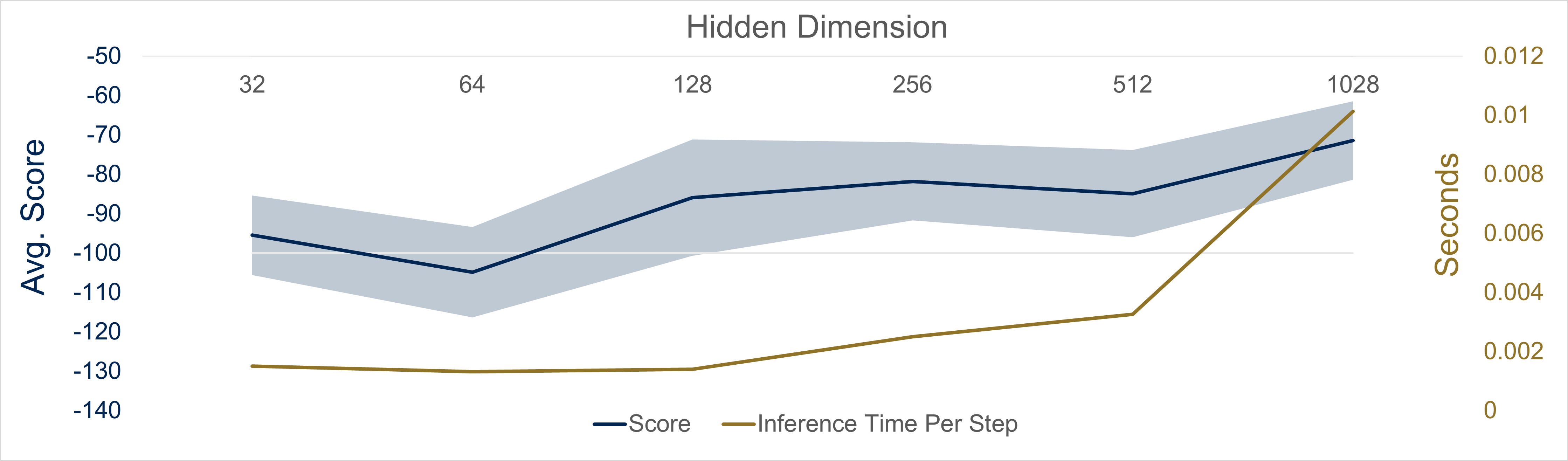}
    \caption{Average score (smaller is better) as more parameters are added to the model (standard deviation plotted as the shaded region).}
    \label{fig:enter-label}
\end{figure}

\textbf{Choice of GNN.} To evaluate the model's sensitivity to the choice of GNN, we evaluate models trained in the CC2 environment with several different base GNNs, namely SAGE~\cite{sage}, GAT~\cite{gat}, GIN~\cite{gin}. 
All training parameters are the same, and models are each trained for 100k episodes. 
We observe that each model performs roughly as well as the others. 
GAT has the best average performance overall, but when the standard error is taken into account, it is only better than SAGE with significance ($p=0.04$, while $p>0.1$ for the other two models). 
These results support our claim that the approach is model-agnostic; our approach is not sensitive to the choice of GNN with statistical significance. 

\begin{table}[!htp]\centering
\caption{Scores of models with different base GNNs on CC2}\label{tab:gnn_ablation}
\scriptsize
\begin{tabular}{lrrrrrrrr}\toprule
& &\multicolumn{2}{c}{30 Steps} &\multicolumn{2}{c}{50 Steps} &\multicolumn{2}{c}{100 Steps} \\\cmidrule{3-8}
&Total &B-Line &Meander &B-Line &Meander &B-Line &Meander\\\midrule
GCN &-80.32 $\pm$ 0.99 &-5.521 &-8.136 &-10.718 &-12.541 &-21.709 &-21.698 \\
SAGE &-84.23 $\pm$ 1.05 &-5.458 &-9.563 &-8.727 &-15.069 &-19.232 &-26.18 \\
GAT &-77.59 $\pm$ 1.39 &-5.164 &-9.404 &-9.373 &-14.282 &-16.937 &-22.434 \\
GIN &-81.44 $\pm$ 1.05 &-5.809 &-8.003 &-9.954 &-12.802 &-22.692 &-22.179 \\
\bottomrule
\end{tabular}
\end{table}

\section{Environment Descriptions}
\label{sec:appendix_env_descriptions}
\textbf{The Yawning Titan environment} represents nodes with two features each: one representing how likely an attacker is to be successful at compromising that node, and one representing whether they have been compromised. 
Additionally, the full list of edges in the graph representing the network is also available. 
We use the combination of node features and edge information as inputs to the GNNs that power our models. 
There are also three actions in this environment
These actions are all listed and described in Table~\ref{tab:yt_actions}. 
In practice, however, we do not implement the Sleep action, as it is always better to Upgrade a node rather than do nothing. 

\begin{table}[h]
    \centering
    \caption{Yawning Titan Actions}
    \label{tab:yt_actions}
    \scriptsize
    \begin{tabular}{lp{5.5cm}p{5.5cm}}
         \toprule
         Action & Description &Effect \\
         \midrule 
         Restore & Removes the attacker from the selected host. Simulates restoring from a previously saved image. & Update node features to ``non-compromised" and reset the ``vulnerability" feature to its initial value \\
         Upgrade & Makes the selected host more difficult to compromise in the future. Simulating a software upgrade or patch. & Decrease the node's ``vulnerability" feature by 0.2.\\
         Sleep & Do nothing \\
         \bottomrule
    \end{tabular} 
\end{table}

The reward function is the ratio of compromised to non-compromised hosts at each time step, with a 100-point bonus for reaching 500 timesteps without allowing every host to be compromised. 
If every host is compromised before step 500, the agent receives a penalty of -100 points. 

\textbf{The CC2 environment} provides highly detailed observations with a great deal of information about each host in the network. 
We only considered the Host, Connection (meaning open ports), File, and Subnet entities when constructing our graphs from the available data. 
Table~\ref{tab:cc2_feats} contains a full description of the features considered for each node type. 
Entries with an asterisk denote engineered features. 
Entries without an asterisk are directly provided in the observations emitted by the environment.  

\begin{table}[t]\centering
\caption{Node features in the CAGE Environments}\label{tab:cc2_feats}
\scriptsize
\begin{tabular}{lllp{8cm}}\toprule
Node Type &Feature &Feature type &Additional information \\\midrule
\multirow{11}{*}{Host} &Architecture &One-hot & \\
&OS Distribution &One-hot & \\
&OS Type &One-hot & \\
&OS Version &One-hot & \\
&OS Kernel Version &One-hot & \\
&OS Patches &One-hot & \\
&Is Critical* &Boolean &Applied to \texttt{Op\_Server0} node in CAGE-2. Means that this node has extra penalties for being compromised \\
&Is User* &Boolean & \\
&Is Server* &Boolean & \\\midrule 
\multirow{10}{*}{Files} &File Type &One-hot & \\
&File Path &One-hot & \\
&Version &One-hot & \\
&Type &One-hot & \\
&Vendor &One-hot & \\
&Density &Float & \\
&Signed &Boolean & \\
&User Permissions &int[0-7] & \\
&Group Permissions &int[0-7] & \\
&Default Permissions &int[0-7] & \\\midrule 
Connection &Is Decoy* &Boolean &If the connection is to a decoy process running on the host \\\midrule 
Subnet &None & &Structural node to connect hosts residing in the same subnet \\
\bottomrule\\
\multicolumn{4}{l}{*~Engineered features that are not provided by default}
\end{tabular}
\end{table}

The environment allows for ten total action types, each described in greater detail in Table~\ref{tab:cage_actions}.  
Unlike the Yawning Titan environment, in CC2, actions may have a cost associated with them. 
In this environment, only the Restore action, which fully wipes a host and restores it from an image, has a cost of -1. 
The Analyze action provides more information about the files on a given host. 
The Remove action kills any user-level shells on the host. 
Finally there are seven different Decoy actions, each starting a honeypot process on a different port. 
Certain decoys are only allowed on specific operating systems, and decoys can only be created if the host is not running any other processes on the port they would occupy. 

\begin{table}[htbp]
    \centering
    \scriptsize
    \caption{Blue agent action space in the CAGE environments}
    \label{tab:cage_actions}
    \begin{tabular}{lp{5.5cm}p{5.5cm}r}
        \toprule 
        Action &Description &Effect &Cost \\\midrule
        Monitor & Review network traffic logs for suspicious activity. This action is taken implicitly every turn, but if selected explicitly, it functions as a no-op action. & Create edges from host nodes to connection nodes if network activity is observed. & 0\\
        Analyze & Attempt to learn it if this host has been compromised by the red agent. &Update node features if it is compromised. Add file nodes and edges connecting them to host if found during scan. &0 \\[2em]
        Remove & Attempt to remove the red agent from this machine. However, if the red agent has privilege escalated to root, this action will fail. &Update node features if successful. &0\\[2em]
        Restore & Guarantees the red agent will be removed from this host, but causes significant disruption to the network & Remove all edges to open port nodes adjacent to the host. Reset the host node's features to show it as not compromised. &$1$ \\[2em]
        Decoy & Open a port on this host to act as a honeypot for the red agent. If the red agent attempts to compromise that port, it will fail. There are 7 types of decoy actions. Each may only be used if the port it would open is not in use, and sometimes only if the host uses a specific OS. & Create a new open port node with an edge to connecting it to the host. &0\\
        AllowTraffic** &Modifies a firewall rule to allow communication between two subnets &Creates an edge between the two subnet nodes &0 \\
        BlockTraffic** &Modifies a firewall rule to disallow communication between two subnets &Deletes an edge between the two subnet nodes &0 \\
        \bottomrule \\
        \multicolumn{4}{l}{**~Actions only available in the CC4 environment.} \\
    \end{tabular}
\end{table}

The reward function for this environment is the number of compromised hosts in the network. 
For each user-type host compromised, a penalty of 0.1 points is applied; compromised server-type hosts are penalized 1.0 points; if the critical asset \texttt{OpServer0} is compromised, there is a penalty of 10 points. 
The full reward function is the sum of penalties from host compromises and the cost of the action selected by the agent. 

\textbf{The CC4 Environment} is similar to the CC2 environment. 
Observations are identical between the environments with two differences. 
In CC4, as timesteps progress, the environment goes through three distinct phases. 
During each phase, there are different firewall rules and communication policies that are allowed. 
We communicate the current phase to the agent as input to the global vector network. 
The second difference is that agents are each allowed to communicate 8 bits to one another. 
We used this to indicate the state of each subnet the agent was defending. 
For each subnet, the first bit represented if any host in the subnet had been compromised. 
We used the second bit as a check bit to determine the difference between no message (0,0) and a message of no compromise (0,1). 
These messages were used as features for the subnet nodes they corresponded to. 
Another important distinction is that agents do not have full knowledge of the network. 
Each agent only receives observations about the subnets it defends; their only knowledge of subnets outside of their influence is through the messages sent by other agents. 

Last, the reward structure in CC4 is different from CC2. 
Rewards are based on the ability of green agents (representing employees) to do either local work, or access services throughout the network. 
The red agent attempts to disrupt hosts, and if it is successful, when a green action tries to use it, it will fail, and the blue agent will be penalized. 
The exact penalty amount changes based on the current phase and the subnet of the host the green agent couldn't access.

\end{document}